\documentclass[10pt,twocolumn,letterpaper]{article}

\usepackage[pagenumbers]{cvpr} 

\usepackage{graphicx}
\usepackage{amsmath}
\usepackage{amssymb}
\usepackage{booktabs}

\usepackage[pagebackref,breaklinks,colorlinks]{hyperref}

\usepackage[capitalize]{cleveref}
\crefname{section}{Sec.}{Secs.}
\Crefname{section}{Section}{Sections}
\Crefname{table}{Table}{Tables}
\crefname{table}{Tab.}{Tabs.}

\def\cvprPaperID{398}
\def\confName{CVPR}
\def\confYear{2023}

\usepackage[utf8]{inputenc} 
\usepackage{url}            
\usepackage{booktabs}       
\usepackage{amsfonts}       
\usepackage{nicefrac}       
\usepackage{microtype}       
\usepackage{amsmath}
\usepackage{cite}
\usepackage{enumitem}
\usepackage{xspace}
\usepackage{color}
\usepackage{graphicx}
\usepackage{textcomp,gensymb}
\usepackage{epsfig}
\usepackage{subcaption}
\usepackage[table,xcdraw]{xcolor}
\usepackage{multirow}
\usepackage{arydshln}

\usepackage{stackengine}
\newcommand\xrowht[2][0]{\addstackgap[.5\dimexpr#2\relax]{\vphantom{#1}}}
\usepackage[noend]{algpseudocode}
\algrenewcommand\algorithmicrequire{\textbf{Input:}}

\newcommand{\green}[1]{{\color{green}#1}}

\newcommand{\fig}{Figure~}
\newcommand{\figs}{Figures~}
\graphicspath{{./figure/}}

\newcommand{\uc}{{\small \sf \mbox{UpCycling}}\xspace}  
\newcommand{\ucr}{{\small \sf \mbox{UpC-R}}\xspace} 
\newcommand{\fgt}{{\textit{\mbox{F-GT}}}\xspace}

\newcommand{\av}{\mbox{AV}\xspace}
\newcommand{\avs}{\mbox{AVs}\xspace}
\newcommand{\ddd}{{3D object detection}\xspace}
\begin{document}
\title{UpCycling: Semi-supervised 3D Object Detection without Sharing\\ Raw-level Unlabeled Scenes}

\author{
Sunwook Hwang$\dagger$\ \ \ \
Youngseok Kim$\dagger$\ \ \ \
Seongwon Kim$\S$\ \ \ \ \\
Saewoong Bahk$\dagger$ $\ast$\ \
Hyung-Sin Kim$\ddagger$ $\ast$\\
{\small $\dagger$Department of Electrical and Computer Engineering, Seoul National University, $\ast$Corresponding author}\\
{\small $\S$Multimodal AI in SK Telecom, Seoul, Korea, $\ddagger$Graduate School of Data Science, Seoul National University}\\
{\tt\small 
\{swhwang, yskim\}@netlab.snu.ac.kr, 
s1kim@sk.com,
\{sbahk, hyungkim\}@snu.ac.kr
}
}

\maketitle

\begin{abstract}\label{sec:abstract}
  Semi-supervised Learning (SSL) has received increasing attention in autonomous driving to reduce the enormous burden of 3D annotation.
  In this paper, we propose \uc, a novel SSL framework for 3D object detection with zero additional raw-level point cloud: learning from unlabeled de-identified intermediate features (\ie, ``smashed'' data) to preserve privacy.
  Since these intermediate features are naturally produced by the inference pipeline, no additional computation is required on autonomous vehicles.
  However, generating effective consistency loss for unlabeled feature-level scene turns out to be a critical challenge. 
  The latest SSL frameworks for 3D object detection that enforce consistency regularization between different augmentations of an unlabeled raw-point scene become detrimental when applied to intermediate features.
  To solve the problem, we introduce a novel combination of \textit{hybrid pseudo labels} and \textit{feature-level Ground Truth sampling} (\fgt), which safely augments unlabeled multi-type 3D scene features and provides high-quality supervision.
  We implement \uc on two representative \ddd models: SECOND-IoU and PV-RCNN.
  Experiments on widely-used datasets (Waymo, KITTI, and Lyft)  verify that \uc outperforms other augmentation methods applied at the feature level. In addition, while preserving privacy, \uc performs better or comparably to the state-of-the-art methods that utilize raw-level unlabeled data in both domain adaptation and partial-label scenarios.
\end{abstract}
\vspace{-15pt}
\section{Introduction}\label{sec:intro}
\vspace{-2pt}

Although the concept of Autonomous Vehicles (\avs) has been around for years,
ensuring the safety of users driving \avs on real roads via 3D object detection models is still challenging.
To this end, there have been continuous efforts to collect large datasets of 3D road scenes
and annotate them carefully~\cite{kitti2012CVPR,waymopaper2020cvpr,lyft_dataset}.
While rapid advances in sensor technology facilitate the collection of 3D scenes at scale,
the severe \textit{annotation burden} remains as a main challenge.
To alleviate the problem, a couple of semi-supervised learning (SSL) methods for \ddd have been proposed recently, such as a combination of perturbation and consistency loss~\cite{sess2020cvpr} and confidence-based filtering using IoU prediction results~\cite{3dioumatch2021cvpr}.
However, these methods learn from unlabeled raw 3D scenes. Collecting a vast amount of raw-level road scenes from \avs can potentially cause disclosure of sensitive private information on the roads~\cite{xiong2020edge,privacy14secpri,e_pp_v2x2020sensors}.

\begin{figure}[t]
  \centering
  \begin{subfigure}[t]{.49\columnwidth}
    \centering
    \includegraphics[width=.75\columnwidth]{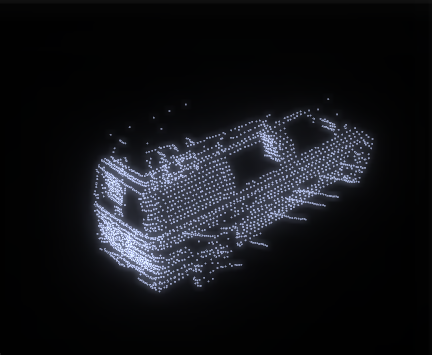}
    \caption{Raw-point data}
  \end{subfigure}
  \hfill
  \begin{subfigure}[t]{.49\columnwidth}
    \centering
    \includegraphics[width=.75\columnwidth]{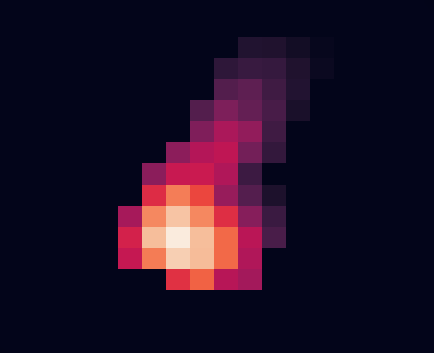}
    \caption{Feature data produced from the 3D object detection network}
  \end{subfigure}
  \hfill
  \begin{subfigure}[t]{.49\columnwidth}
    \centering
    \includegraphics[width=.99\columnwidth]{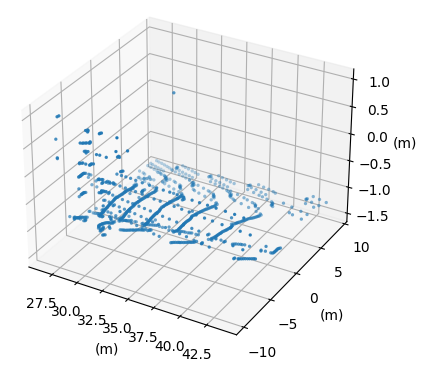}
    \caption{Original point cloud scene}
  \end{subfigure}
  \hfill
  \begin{subfigure}[t]{.49\columnwidth}
    \centering
    \includegraphics[width=.99\columnwidth]{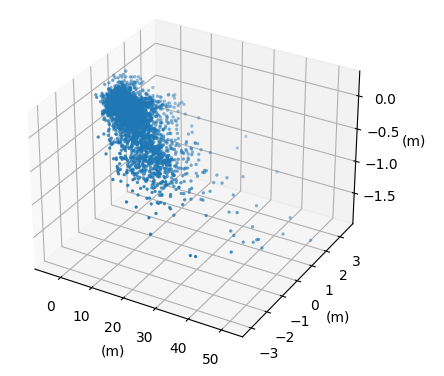}
    \caption{Restored point cloud scene using the inversion attack}
  \end{subfigure}
  \vspace{-7pt}
  \caption{Visualization of point cloud scenes. \uc improves level of privacy protection since an original point cloud scene cannot be restored from its intermediate feature.}
  \label{fig:motivation}
  \vspace{-10pt}
\end{figure}

Given that the problem of potential \textit{privacy leakage} from raw data collection exists in various applications, a number of studies have tried to not deal with raw data directly.
Going beyond encrypting raw data~\cite{xiong2020edge}, federated learning~\cite{Konecny2017arxiv,gupta2018distributed} makes each edge node consume its data locally to train the model and share the model weights (or gradients) instead of raw data. Split learning~\cite{gupta2018distributed,vepakomma2018split,singh2019detailed} designs edge nodes to not share raw data but its intermediate feature (\ie, smashed data) that comes from passing through early-stage layers of the model.
However, these approaches require local training~\cite{fedpaq2020pmlr,uveqfed2021tsp}, which makes resource-constrained \avs suffer more \textit{computation overhead}. Given that \avs use significant computing resources to process inference pipelines for 3D detection during driving, such additional computation hinders continuous model updates in natural driving conditions.

In this paper, we aim to address all the three issues: labeling cost, privacy, and \av-side computation overhead.
To ensure this end, we propose \uc, a novel SSL framework that does not utilize unlabeled raw 3D scenes (\fig\ref{fig:motivation}(a)) but
\textit{de-identified, unlabeled} intermediate features (\fig\ref{fig:motivation}(b)) to advance 3D object detection models.
Since an unlabeled intermediate feature is naturally produced during a regular detection pipeline with the 3D scene, \uc requires neither additional \av-side computation (\eg, local training) nor server-side annotation burden.
Further, sharing features instead of raw 3D scenes improves the level of privacy protection as the detection pipeline includes nonlinear layers and compression~\cite{voxelnet2018cvpr,second2018sensors,pointpillar2019cvpr,voxelrcnn2021aaai,pvrcnn2020cvpr}.
Because the process in the nonlinear layers~\cite{vepakomma2018no} is irreversible, the original scene cannot be completely restored from its intermediate feature.
As depicted in \figs\ref{fig:motivation}(c) and (d), the inversion attack~\cite{Dosovitskiy_2016_CVPR} attempted on the server side to restore the raw-point data does not result in a successful restoration.\footnote{For further details, please refer to Section~\ref{subsec:privacy} and Supplementary material where more comprehensive information is provided.}

To realize the advantages, \uc should provide an effective feature-based SSL method for 3D object detection, which involves two challenges:
(1) augmenting unlabeled intermediate features reliably to increase data diversity~\cite{csd19nips,Laine2017iclr} and
(2) providing high-quality pseudo labels to supervise these augmented features.
The state-of-the-art (SOTA) semi-supervised 3D object detection frameworks~\cite{sess2020cvpr,3dioumatch2021cvpr} generate consistency loss between weak and strong augmentations of a 3D point scene. However, the augmentation methods targeting raw-level point clouds become detrimental when applied at a feature level. This is because an intermediate feature is a smashed form of its original 3D scene and has multiple types depending on the \ddd models, such as grid- and set-types.
Therefore, naïve application of the point augmentation methods at a feature level damages the important information in the 3D scene, which causes the pseudo labels to suffer from significant noise.

\begin{figure*}[t]
\vspace{-1ex}
  \centering
  \includegraphics[width=0.88\textwidth]{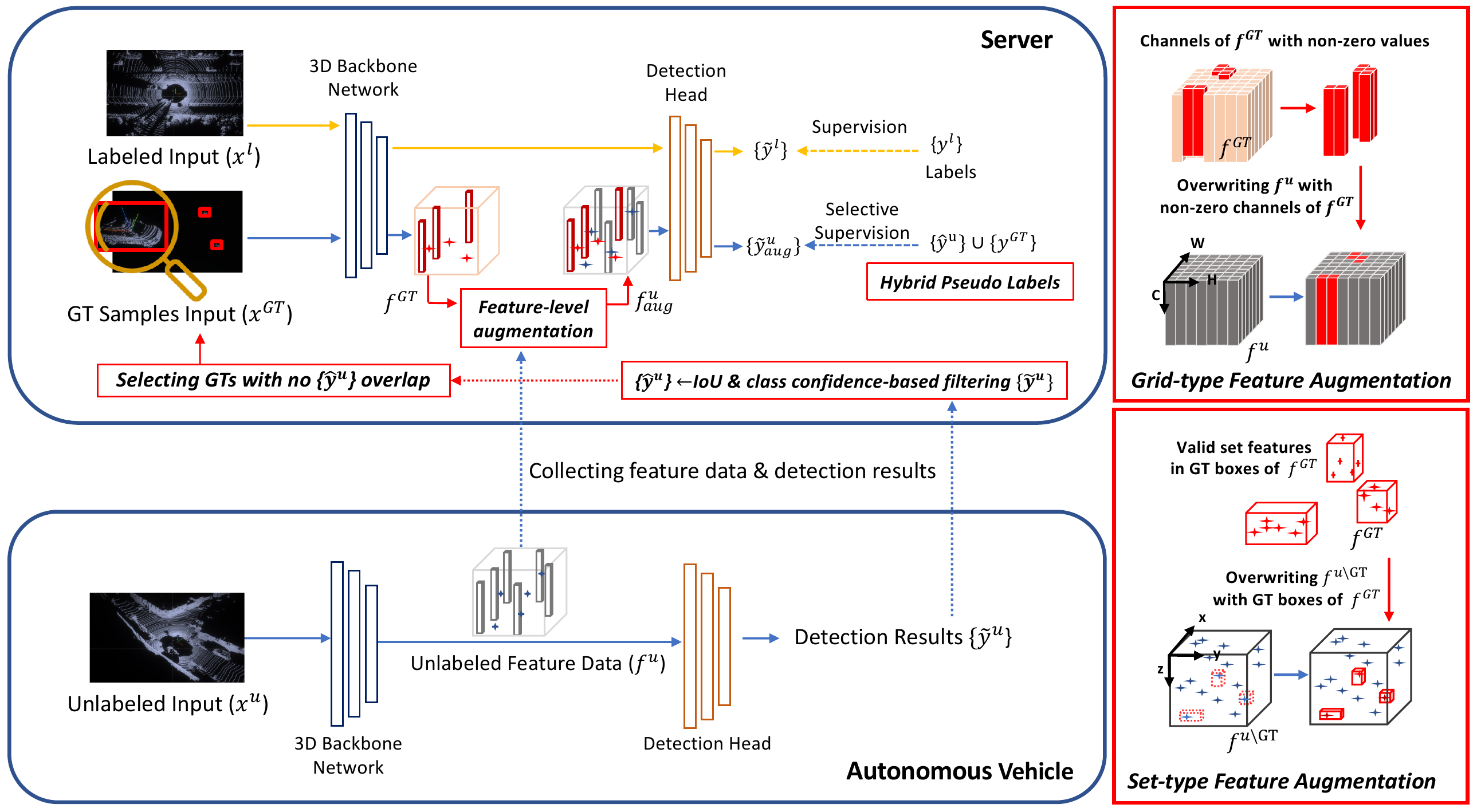}
  \vspace{-8pt}
  \caption{
  Overview of the \uc framework.
  $\mathbf{f}^u$ and $\{{\mathbf{\tilde y}}^u\}$ refer to unlabeled feature data and detection results from \avs, respectively.
  IoU and class confidence-based threshold filters detection results to obtain $\{\mathbf{\hat y}^u\}$.
  GTs that do not overlap with $\{\mathbf{\hat y}^u\}$ are sampled to form high-quality hybrid pseudo labels.
  To obtain data diversity, \uc augments the collected unlabeled feature-level data $\mathbf{f}^u$ with GT sampling (\fgt).
  The resulting augmented feature, $\mathbf{f}^u_{aug}$, is supervised by the high-quality hybrid pseudo labels.
  }
  \vspace{-12pt}
  \label{fig:overview}
\end{figure*}

To address the challenges, we propose high-quality \textit{hybrid pseudo labels} and feature-level ground-truth sampling (\fgt). Combining these methods not only achieves significant data diversity but also improves quality of pseudo labels by adding zero-noise labels.
We implement \uc on two representative 3D detection models, PV-RCNN~\cite{pvrcnn2020cvpr} and SECOND-IoU~\cite{openpcdet2020},\footnote{SECOND-IoU adds an IoU module to the original SECOND model~\cite{second2018sensors}.}
and perform various experiments on three major datasets for \av applications, KITTI~\cite{kitti2012CVPR}, Lyft~\cite{lyft_dataset}, and Waymo~\cite{waymopaper2020cvpr}.
The results demonstrate the effectiveness of \uc in both partial-label and domain adaptation scenarios.

The contributions of this work are summarized as follows:
\vspace{-4ex}
\begin{itemize}[leftmargin=*]
  \item \uc is the first framework that tackles labeling cost, privacy leakage, and \av-side computation cost altogether to train a 3D object detection model, which deeply investigates how to learn from unlabeled intermediate features.
  \vspace{-4ex}
  \item \uc provides a fresh eye on GT sampling in the context of SSL
        since it safely improves data diversity of unlabeled feature-level 3D scenes and  significantly improves pseudo-label quality by providing zero-noise labels.
  \vspace{-3ex}
  \item \uc not only protects privacy but also achieves SOTA accuracy in both domain adaptation and partial-label scenarios, on representative models and datasets for 3D object detection.
\end{itemize}
\vspace{-2ex}
\section{Related Work}\label{sec:related}
\noindent
\textbf{Semi-supervised learning.}
SSL has been actively studied in the context of image classification~\cite{Laine2017iclr,Tarvainen2017nips,sohn2020fixmatch,Miyato2019tpami}. Most of the recent SSL methods~\cite{csd19nips,Laine2017iclr,Tarvainen2017nips,Miyato2019tpami} leverage consistency regularization which trains the model to obtain consistent prediction results across label-preserving data augmentation.
In the SSL frameworks, proper data augmentation is essential, which should significantly increase diversity effect without losing consistency with the original data~\cite{cubuk2020randaugment,devries2017improved}.
Accurate pseudo-labeling is another crucial element for SSL to provide high-quality supervision for unlabeled data~\cite{lee2013pseudo,sohn2020fixmatch}.
While there have been only a couple of studies on SSL for 3D object detection~\cite{sess2020cvpr,3dioumatch2021cvpr}, data augmentation and pseudo-labeling are still important. SESS~\cite{sess2020cvpr} targets indoor 3D object detection, leveraging a teacher-student architecture that takes differently augmented 3D scenes as inputs and utilizes three kinds of consistency losses between outputs. 3DIoUMatch~\cite{3dioumatch2021cvpr} improves quality of pseudo labels with confidence-based filtering in the IoU-guided NMS stage.
However, the SSL methods require direct access to a vast amount of raw data, which causes potential privacy leakage.

\vspace{1.5pt}
\noindent
\textbf{Feature-level data augmentation.}
Data diversity can be limited when augmenting only raw data. To further increase diversity, feature-level data augmentation has been investigated~\cite{li2021simple,chu2020feature,cen2021deep,Liu_2018_CVPR,pmlr-v97-verma19a}.
In image classification tasks, adding Gaussian noise to feature-level data gains more data diversity for training and domain generalization~\cite{li2021simple}.
The work in~\cite{chu2020feature,cen2021deep,Liu_2018_CVPR} resolves lack of data for specific classes by using feature augmentation.
Feature augmentation is also applied to few-shot learning in NLP tasks~\cite{kumar2019closer}.
To our knowledge, however, feature-level augmentation has not been studied in the context of semi-supervised \ddd.

\vspace{1.5pt}
\noindent
\textbf{Private representation learning.}
Private representation learning~\cite{Konecny2017arxiv,gupta2018distributed} aims to learn from various clients without sharing their raw data, which heavily relies on local training at resource-constrained clients.
Federated learning designs clients to not share any data but model weights or gradients with the server. Due to the local computation burden for training the whole model, federated learning methods~\cite{fl2017pmlr,reddi2020adaptive,wang2020federated} face significant hurdles in training large neural nets.
Split learning~\cite{gupta2018distributed,vepakomma2018split,singh2019detailed} is more similar to \uc in that clients share intermediate features of local data with the server. However, it still requires local training of early layers of the model. Continuous communication burden during training is another problem of these approaches.

\vspace{1pt}
\noindent
\textbf{\ddd models.}
Main challenges in 3D object detection come from the irregular and sparse positions of 3D point clouds.
To address the issues, some researches~\cite{pointnet2017cvpr,pointrcnn2019CVPR} opt for point-based methods that extract set-type features by processing raw point clouds directly~\cite{qi2017pointnet++}.
Other approaches~\cite{second2018sensors,voxelnet2018cvpr,pointpillar2019cvpr,pvrcnn2020cvpr} suggest voxel-based methods, which first voxelize a point cloud and extract grid-type features with 3D convolution networks.
Therefore, \uc should be able to handle both grid- and set-type unlabeled features. Specifically, we adopt two representative 3D object detectors: voxel-based SECOND-IoU~\cite{second2018sensors,openpcdet2020}
and PV-RCNN~\cite{pvrcnn2020cvpr} that mixes point- and voxel-based methods.

\vspace{-0.5ex}
\section{Method}\label{sec:method}
\vspace{-0.5ex}

\subsection{Problem Definition}
Given a 3D point cloud scene $\mathbf{x}$, we aim to detect a set of 3D bounding boxes and class labels for all objects in $\mathbf{x}$, denoted as $\{\mathbf{y}\}$. We perform this task under a new challenging SSL scenario with unlabeled de-identified data: in contrast to the regular SSL setting, unlabeled raw-level point clouds are not available. Specifically, we have access to $N$ training samples, including $N^l$ labeled point clouds $\{\mathbf{x}^l_i,\{\mathbf{y}^l_i\}\}^{N^l}_{i=1}$ and $N^u$ unlabeled scenes in the form of \textit{intermediate feature} $\{\mathbf{f}^u_i\}^{N^u}_{i=1}$. Here $\mathbf{f}^u$ is the output of the backbone network for an unlabeled point cloud $\mathbf{x}^u$.

\vspace{-0.5ex}
\subsection{UpCycling Framework}
\vspace{-0.8ex}

\fig\ref{fig:overview} depicts the overall \uc framework incorporating server- and \av-side operations.
For initialization, the server trains a \ddd model on its labeled data $\{\mathbf{x}^l_i,\{\mathbf{y}^l_i\}\}^{N^l}_{i=1}$ and shares the pre-trained model with \avs.
\uc targets the latest 3D detection models with an IoU module that returns \textit{confidence scores} for bounding box localization. In this paper, we apply \uc in PV-RCNN~\cite{pvrcnn2020cvpr} and SECOND-IoU~\cite{openpcdet2020}.
PV-RCNN is the representative IoU-aware model for \ddd and SECOND-IoU is a modified version of SECOND~\cite{second2018sensors} with addition of IoU module.\\
\indent For autonomous driving, \avs continuously perform the model's detection pipeline for newly observed 3D scenes.
At the same time, to further update the model with more 3D scenes in diverse environments, each \av sends a new 3D scene $\mathbf{x}^u$'s intermediate feature $\mathbf{f}^u$ to the server, which serves as \textit{de-identified unlabeled training data}. It is noteworthy that \textit{zero additional computation} is needed for the de-identification since the feature naturally comes from processing the 3D backbone network in the detection pipeline.
Each \av also sends the detection results $\{{\mathbf{\tilde y}}^u\}$ to the server.

With the received features and detection results $\{\mathbf{f}^u_i,\{\mathbf{\tilde y}^u_i\}\}^{N^u}_{i=1}$,
the server generates consistency loss in a different way of the SOTA SSL methods on 3D object detection that utilize unlabeled raw-point scenes $\{\mathbf{x}^u_i\}^{N^u}_{i=1}$~\cite{sess2020cvpr,3dioumatch2021cvpr}.
Specifically, given that supervising $\mathbf{f}^u$ by using its detection result $\{\mathbf{\tilde y}^u\}$ again is meaningless, (1) proper augmentation of $\mathbf{f}^u$ and (2) high-quality pseudo labels are essential.

The SOTA methods on semi-supervised 3D object detection~\cite{sess2020cvpr,3dioumatch2021cvpr} take a teacher-student architecture~\cite{Tarvainen2017nips} by using random sampling (RS) for weak augmentation and both RS and Flip for strong augmentation of a point cloud. However, in our scenario where an input is an intermediate feature, the augmentation methods significantly damage the original scene.
Instead, we propose feature-level ground-truth sampling (\fgt) for feature augmentation, as illustrated in Figure~\ref{fig:overview}.
Although ground-truth (GT) sampling has been used as a point cloud augmentation method for supervised 3D object detection~\cite{voxelnet2018cvpr,second2018sensors,pvrcnn2020cvpr,pointpillar2019cvpr,voxelrcnn2021aaai} and is known to provide at most fair performance improvement~\cite{hahner2020quantifying}, we claim that its impact can be more significant when it comes to \textit{feature-level} augmentation of an \textit{unlabeled} 3D scene. This is because \fgt tackles one of the most crucial issues for successful SSL: improving the quality of pseudo labels for unlabeled features by generating \textit{hybrid pseudo labels}.


\subsection{Hybrid Pseudo Labels}
For effective SSL, we adopt \fgt to augment an unlabeled scene feature $\mathbf{f}^u$ and include the sampled GT labels (zero-noise labels) in the pseudo-label set for the unlabeled feature.
By doing so, \uc constructs high quality \textit{hybrid pseudo labels}.

\vspace{2pt}
\noindent
\textbf{Confidence-based pseudo-label filtering.}
First, inspired by 3DIoUMatch~\cite{3dioumatch2021cvpr}, \uc screens the received detection results $\{\mathbf{\tilde y}^u\}$ by using each $\mathbf{\tilde y}^u$'s confidence scores for both object classification and bounding box localization. Assume that ${\tau}_{IoU}$ and ${\tau}_{cls}$ are thresholds for box localization and object classification, respectively. \uc filters out a detection result if its class confidence or localization confidence is lower than the given threshold, leaving a set of high-quality pseudo labels, denoted as $\{\mathbf{\hat y}^u\}$.
The confidence-based pseudo-label filtering is applied for more accurate supervision.

\vspace{2pt}
\noindent
\textbf{Pseudo-label-aware GT sampling.}
When GT sampling is applied for supervised learning, it first constructs a GT database that consists of labeled 3D bounding boxes and point clouds in the boxes, collected from the entire labeled training set $\{\mathbf{x}^l_i,\{\mathbf{y}^l_i\}\}^{N^l}_{i=1}$. To augment a labeled 3D scene $\mathbf{x}^l$, GTs are sampled from the database and randomly placed in the 3D scene. To avoid tampering with GT information, a GT sample that overlaps with a ground-truth bounding box in the original labeled scene is removed.

In contrast, our \fgt aims to augment an \textit{unlabeled} 3D scene feature $\mathbf{f}^u$ without accurate box labels. Instead, given that a set of high-quality pseudo labels $\{\mathbf{\hat y}^u\}$ is provided, \fgt samples GTs that do not overlap with the \textit{pseudo labels}. Importantly, although the pseudo labels are filtered with the two thresholds $\tau_{IoU}$ and $\tau_{cls}$, these thresholds are set moderately~\cite{3dioumatch2021cvpr}, enabling the pseudo labels to cover most objects in the original scene $\mathbf{x}^u$; GT samples are likely to be placed on the background of $\mathbf{x}^u$.

\vspace{2pt}
\noindent
\textbf{Hybrid pseudo-labels.}
To generate pseudo labels that supervise an augmented unlabeled feature $\mathbf{f}^u_{aug}$,
\uc merges the high-quality pseudo-label set for the original feature $\mathbf{f}^u$, $\{\mathbf{\hat y}^u\}$, with the label set for the GT samples, $\{\mathbf{y}^{GT}\}$, resulting in a set of \textit{hybrid pseudo labels} $\{\mathbf{\hat y}^u\}\cup\{\mathbf{y}^{GT}\}$. Given that $\{\mathbf{y}^{GT}\}$ are literally ground-truth labels with \textit{zero noise}, adding these labels to the pseudo labels enables powerful supervision.
Furthermore, generating the hybrid pseudo labels does not need to execute the inference pipeline at the server, since all GT labels are already given.
%
%
\vspace{-0.5ex}
\subsection{Feature-level 3D Scene Augmentation}
\vspace{-0.5ex}

Regarding \fgt, since the server does not have an original unlabeled scene $\mathbf{x}^u$ but only its intermediate feature $\mathbf{f}^u$, it is impossible to directly place GT samples on the point cloud scene. Instead, \fgt generates a separate point cloud input that comprises only GT samples. The GT-only point cloud passes through the model's 3D backbone network, resulting in a GT-only feature $\mathbf{f}^{GT}$.
Note that while the 3D backbone of SECOND-IoU generates only grid-type features, that of PV-RCNN~\cite{pvrcnn2020cvpr} generates both grid- and set-type features. To this end, \fgt augments $\mathbf{f}^u$, grid- or set-type feature, as follow:

\vspace{2pt}
\noindent
\textbf{Grid-type feature augmentation.}
As shown in \fig\ref{fig:overview}, when $\mathbf{f}^u$ and $\mathbf{f}^{GT}$ are grid-type features, \fgt generates an augmented feature by overwriting $\mathbf{f}^u$ with $\mathbf{f}^{GT}$; if a channel on $\mathbf{f}^{GT}$ has non-zero values, the $\mathbf{f}^{GT}$ channel replaces that in $\mathbf{f}^u$.
Giving higher priority for $\mathbf{f}^{GT}$ removes some information included in $\mathbf{f}^u$. However, given that the GT samples take up a tiny portion of an entire scene (\ie, most values in $\mathbf{f}^{GT}$ are zero), only a small number of values in $\mathbf{f}^u$ are modified. In addition, the removed information in $\mathbf{f}^u$ is related to the background since the sampled GTs are not overlapped with pseudo labels, which does not harm model training.

\vspace{2pt}
\noindent
\textbf{Set-type feature augmentation.}
When an unlabeled feature $\mathbf{f}^{u}$ and a GT sample feature $\mathbf{f}^{GT}$ are set types, each of them consists of $n$ represented points, denoted as $\mathbf{f}^{u}=\{f^{u}_{i}\}^n_{i=1}$
and $\mathbf{f}^{GT}=\{f^{GT}_{i}\}^n_{i=1}$, respectively.
In this case, as illustrated in \fig\ref{fig:overview}, \fgt generates an augmented feature as a point set, denoted as $\mathbf{f}^{u}_{aug}=\{f^{u}_{aug,i}\}^n_{i=1}$. To this end, we first exclude the scene feature points $f^u_i$ that are in the GT boxes, generating $\mathbf{f}^{u \backslash GT}$. Then each feature point $f^{u}_{aug,i}$ is randomly sampled from either $\mathbf{f}^{u \backslash GT}$ or $\mathbf{f}^{GT}$.

In doing so, it is important that the scene feature contains much more information than the GT feature; for reasonable augmentation, $\mathbf{f}^{u}_{aug}$ should include scene feature points more than GT feature points. To determine proper sampling frequency, we utilize the information in the grid-type feature that is generated simultaneously with the set-type feature by the 3D backbone network: how many values in the grid-type feature for the scene and GT samples are non-zero.
For example, if the number of grid with non-zero values in the scene and GT features (grid types) is 2000 and 50, respectively, points in the augmented feature set $\mathbf{f}^{u}_{aug}$ is sampled from $\mathbf{f}^{u \backslash GT}$ 400 times more than $\mathbf{f}^{GT}$.

\begin{figure}[t]
  \centering
  \includegraphics[width=0.95\columnwidth]{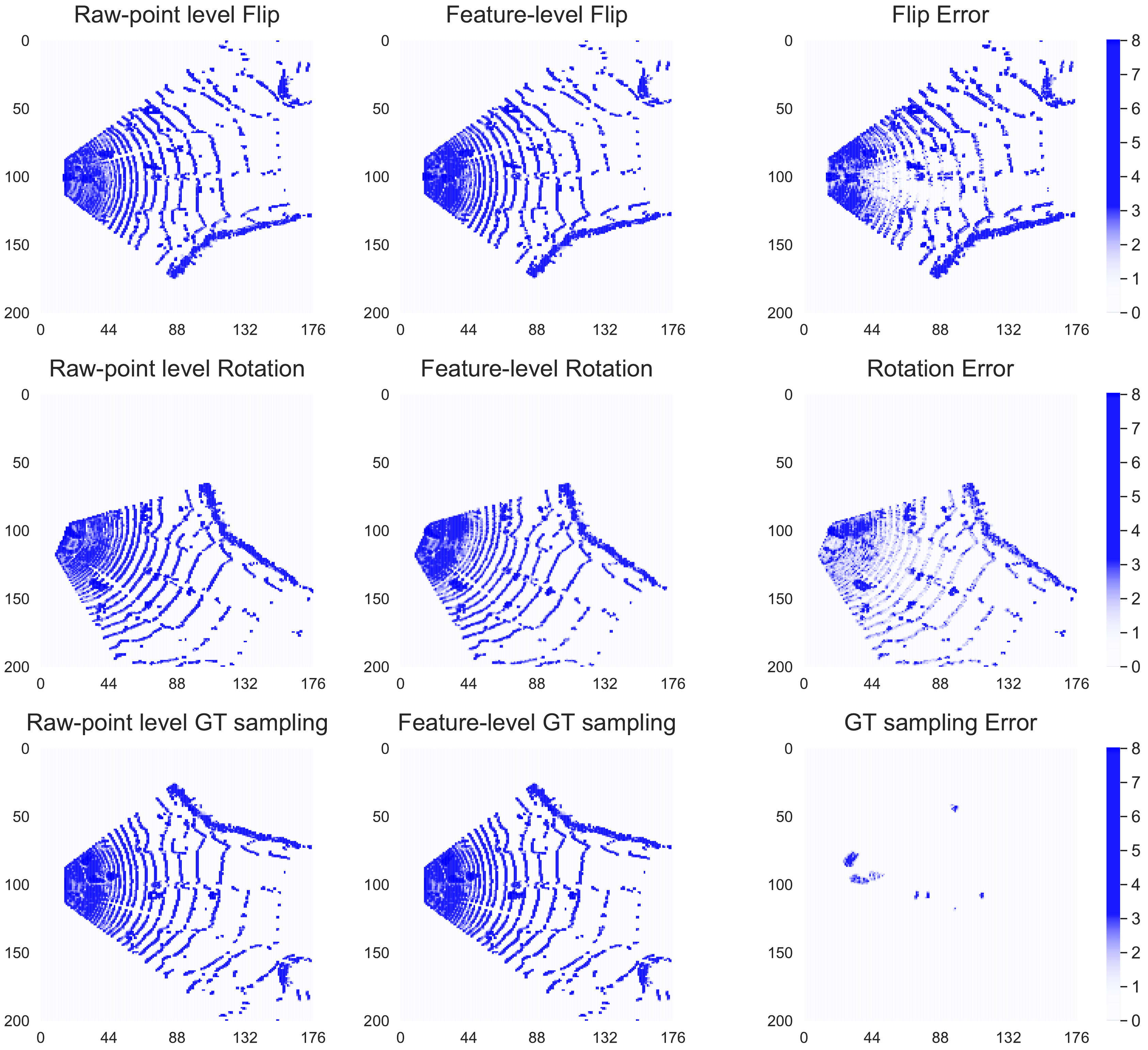}
  \vspace{-1ex}
  \caption{
    Feature-level scenes for three data augmentation methods: Flip (1st row), Rotation (2nd row), and GT sampling (3rd row). Feature-level scenes of raw-point level augmentation are on the left. Feature-level scenes of feature-level augmentation are in the middle.
    Heatmaps of RMSE based on comparison between raw-level and feature-level augmentation scenes are on the right.}
  \label{fig:heatmap}
  \vspace{-12pt}
\end{figure}

\subsection{Loss}

The model's detection head is trained to predict the hybrid pseudo labels for the  augmented feature $\mathbf{f}^u_{aug}$.
Given that our target models have an IoU module as well as a Region Proposal Network (RPN), the unlabeled loss $\mathcal{L}(\mathbf{f}^u_{aug})$ includes loss of each of the two modules as follows:
\begin{equation}\label{eq:unlabel}
  \begin{aligned}
    \mathcal{L}(\mathbf{f}^u_{aug}) = & \mathcal{L}_{loc}^{RPN}(\{\mathbf{\hat y}^u\}\cup\{\mathbf{y}^{GT}\})
    + \mathcal{L}_{loc}^{IoU}(\{\mathbf{\hat y}^u\}\cup\{\mathbf{y}^{GT}\})                                      \\
                                      & + \mathcal{L}_{cls}^{RPN}(\{\mathbf{\hat y}^u\}\cup\{\mathbf{y}^{GT}\}).
  \end{aligned}
\end{equation}
The exact calculation of the three terms depends on the model architecture, following the calculation of supervised loss.
Assuming that a training batch consists of a set of labeled scenes $\{\mathbf{x}^l\}$ and a set of augmented features for unlabeled scenes $\{\mathbf{f}^u_{aug}\}$, the total loss for the batch is calculated as below, where $w$ is the unsupervised loss weight:
\begin{equation} \label{eq:total}
  \mathcal{L}_{total} = \mathcal{L}(\{\mathbf{x}^l\}) + w\mathcal{L}(\{\mathbf{f}^u_{aug}\}).
\end{equation}

\section{Analysis on 3D Scene Feature Augmentation}
\label{sec:preliminary}

In this section, we take a deeper look into subtle feature-level 3D scene augmentation. Specifically, we focus on why widely-used point cloud augmentation methods damage important information when applied at a feature level.

To this end, \fig\ref{fig:heatmap} depicts activation heat maps of the Bird-eye View (BEV) compression module in SECOND-IoU when Flip, Rotation, and GT sampling are applied to an example 3D scene covering $x$, $y$, $z$ axis range 70.4, 80, 4 meters.
The figure shows that in the cases of Flip and Rotation, raw-level augmentation (\ie, flipping/rotating the whole point cloud) and feature-level augmentation (\ie, flipping/rotating the feature vector) result in significantly different activations. In both cases, although the two activation heat maps look similar at a glance, taking the difference between the two causes errors that are widely spread over the entire feature map.
In contrast, when using GT sampling, raw- and feature-level augmentations provide similar activation heat maps. Although some errors exist, they are placed in restricted areas where GT samples are inserted.

\begin{figure}[t]
  \centering
  \includegraphics[width=\columnwidth]{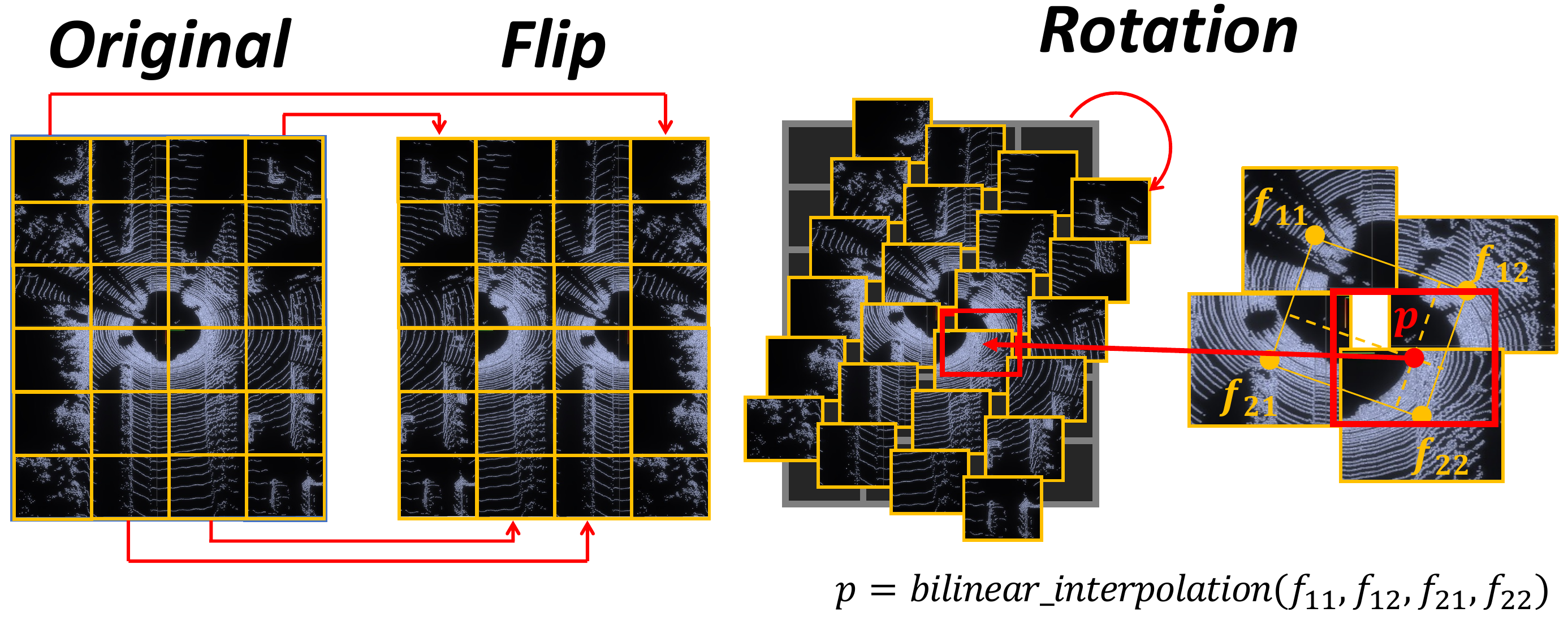}
  \vspace{-20pt}
  \caption{Conceptual images of feature-level augmentation with Flip and Rotation.}
  \vspace{-7pt}
  \label{fig:feature_aug_concept}
\end{figure}

\begin{figure}[t]
  \centering
  \includegraphics[width=.95\columnwidth]{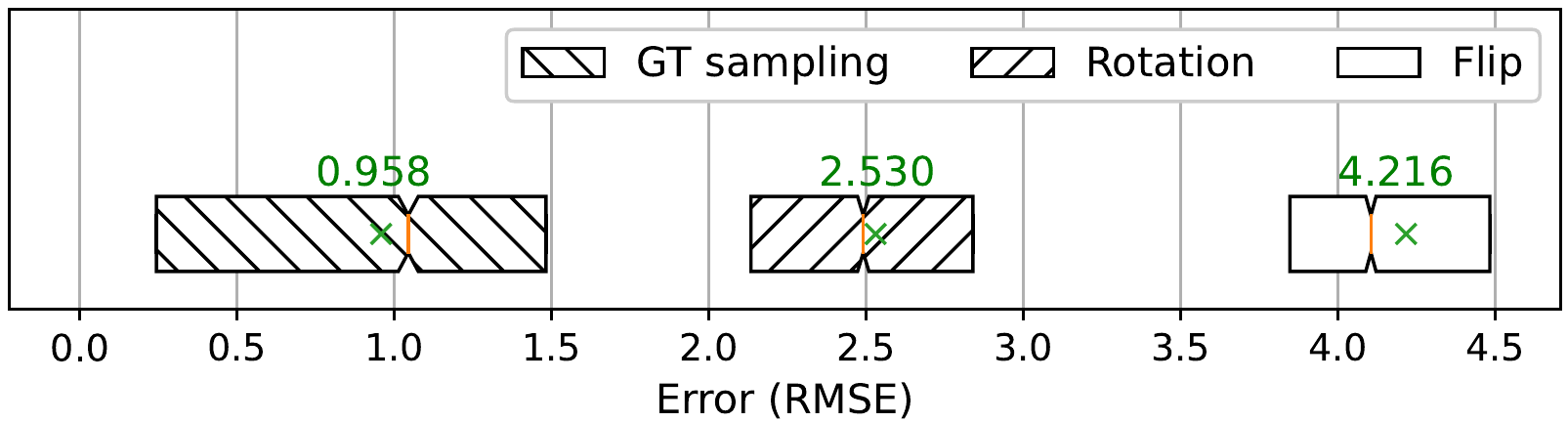}
  \vspace{-5pt}
  \caption{RMSE between raw- and feature-level augmentations of the entire KITTI training dataset. Box range covers the first quartile to the third quartile and the mark `\green{$\times$}' indicates the mean value.}
  \label{subfig:error_bar}
  \vspace{-15pt}
\end{figure}

\fig\ref{fig:feature_aug_concept} provides a visual illustration of Flip and Rotation for feature augmentation.
If a point cloud is voxelized with each voxel producing its feature value, flipping/rotating the feature vector is similar to flipping/rotating voxels.
This means that point locations are shifted not individually but in groups, and the geometric relationship between intra-voxel points is maintained; they are neither flipped nor rotated.
In the worst case, the group (voxel)-wise flipping causes a valid car object to break apart, making its label detrimental to training.
Breaking the geometric relationship between points on the background can also cause severe misinterpretation.
Similarly, the group-wise rotation breaks the geometric relationship mildly and its bilinear interpolation creates the errors, which is not proper for augmentation.

\fig\ref{subfig:error_bar} confirms our description by showing the average of root mean square error (RMSE) between raw- and feature-level augmentations in the KITTI dataset.
This plot illustrates that feature-level Flip and Rotation severely damage the original scene, in contrast to GT sampling, which only produces minor errors.
\vspace{-5pt}
\section{Experiments}
\label{sec:exp}

\subsection{Experimental Setup}
\vspace{-2pt}
\noindent\textbf{Scenarios.}
To demonstrate the effectiveness of \uc in various practical situations, we conduct experiments in both domain adaptation and partial-label scenarios.
The domain adaptation task is to adapt the model, which is trained on abundant labeled data in the source domain, to an unseen target domain that provides only unlabeled data.
In the partial-label scenario, the model is trained and tested in the same domain but most of the training data is unlabeled.

\vspace{1pt}\noindent
\textbf{Datasets.}
We choose three datasets widely used for detection applications of \avs:~Waymo~\cite{waymopaper2020cvpr}, Lyft~\cite{lyft_dataset}, and KITTI~\cite{kitti2012CVPR}.
Among the three, the Waymo dataset is the most diverse and the largest in volume.
The 3D scenes in the Waymo dataset are captured in Phoenix, Mountain View, and San Francisco, the US, under multiple weather and time settings.
The Lyft dataset is collected around Palo Alto, the US, in clear weather in the daytime.
The KITTI dataset is collected in Karlsruhe, Germany, in clear weather during the daytime. Due to regional characteristics, car sizes in KITTI are different from those in Waymo and Lyft~\cite{wang2020train}.
We focus on car objects in this section and more details are in the supplementary material.

\vspace{1pt}\noindent
\textbf{Implementation details.}
When training a model with \uc, we set the two filtering thresholds ${\tau}_{IoU}$ and ${\tau}_{cls}$ to 0.5 and 0.4, respectively, and the weight for the loss $\mathcal{L}(\{f^u_{aug}\})$ is set as $w=1$.
We set the ratio of labeled data to unlabeled data in a mini-batch to 1:2 and 1:1 for domain adaptation and partial-label experiments, respectively. Importantly, \fgt samples GT boxes only from the labeled dataset: the source domain data in the domain adaptation scenario and a small portion of labeled data in the partial-label scenario.
Lastly, \uc freezes the 3D backbone network after training it on the labeled data to prevent the divergence between an intermediate feature from the server's 3D backbone network and that collected from \avs. Therefore, \uc updates only the detection head using unlabeled feature-level data.
More details are in the supplementary material.

\begin{table}[t]
  \centering
  \caption{Effects of feature augmentation methods in a partial-label scenario where the \ddd model is SECOND-IoU and 10\% training data is labeled in KITTI.
  }
  \vspace{-4pt}
  \scalebox{.835}{
    \begin{tabular}{c|ccccc|rrr}
      \multicolumn{1}{l}{\multirow[b]{2}{*}{Policy \#}} & \multicolumn{1}{l}{\multirow[b]{2}{*}{\rotatebox{90}{\textit{Flip}}}} & \multicolumn{1}{l}{\multirow[b]{2}{*}{\rotatebox{90}{\textit{Noise}}}} & \multicolumn{1}{l}{\multirow[b]{2}{*}{\rotatebox{90}{\textit{RS}}}} & \multicolumn{1}{l}{\multirow[b]{2}{*}{\rotatebox{90}{\textit{Rot.}}}} & \multicolumn{1}{l}{\multirow[b]{2}{*}{\rotatebox{90}{\fgt}}} & \multicolumn{3}{c}{$AP_{3D}$}                                                                       \\
      \multicolumn{1}{l}{}                              & \multicolumn{1}{l}{}                                                  & \multicolumn{1}{l}{}                                                   & \multicolumn{1}{l}{}                                                & \multicolumn{1}{l}{}                                                  & \multicolumn{1}{l}{}                                         & Easy                            & Mod                             & Hard                            \\\hline \hline
      Baseline                                          &                                                                       &                                                                        &                                                                     &                                                                       &                                                              & \multicolumn{1}{r}{$70.58$}     & \multicolumn{1}{r}{$56.00$}     & \multicolumn{1}{r}{$47.94$}     \\ \hdashline
      1                                                 & \checkmark                                                            &                                                                        &                                                                     &                                                                       &                                                              & $-16.31$                        & $-20.09$                        & $-19.79$                        \\
      2                                                 &                                                                       & \checkmark                                                             &                                                                     &                                                                       &                                                              & $+0.03$                         & $+0.13$                         & $-1.23$                         \\
      3                                                 &                                                                       &                                                                        & \checkmark                                                          &                                                                       &                                                              & $+2.47$                         & $-0.96$                         & $+0.63$                         \\
      ~~4$\ast$                                         & \checkmark                                                            &                                                                        & \checkmark                                                          &                                                                       &                                                              & $-11.69$                        & $-13.75$                        & $-13.32$                        \\
      5                                                 &                                                                       &                                                                        &                                                                     & \checkmark                                                            &                                                              & $+4.80$                         & $+5.42$                         & $+7.96$                         \\
      \cellcolor[HTML]{EFEFEF}\uc                       &                                                                       &                                                                        &                                                                     &                                                                       & \cellcolor[HTML]{EFEFEF}\checkmark                           & \cellcolor[HTML]{EFEFEF}$+$\textbf{7.81} & \cellcolor[HTML]{EFEFEF}$+$\textbf{7.87} & \cellcolor[HTML]{EFEFEF}$+$\textbf{8.14} \\
    \end{tabular}
  }
  \vspace{-10pt}
  \label{tab:feature_aug_performance}
\end{table}

\subsection{Effect of Feature Augmentation Schemes} \label{subsec:featureaug}

First, we investigate feature augmentation deeply by evaluating the superiority of \fgt, which is utilized for \uc, to other augmentation schemes in a partial-label scenario. To this end, we train SECOND-IoU on the KITTI dataset when only 10\% of its training data is labeled. Importantly, given that the KITTI dataset is originally shuffled regardless of place and time sequence, we rearrange it in chronological order for each place to prevent the data leakage between the labeled and unlabeled sets~\cite{kittisorting}.

\vspace{3pt}
\noindent
\textbf{Comparison schemes.}
In this scenario, \textbf{Baseline} trains the model using only the limited amount of labeled data.
\textbf{Flip} and \textbf{RS} are used in the SOTA SSL methods on 3D object detection to augment raw-level 3D scenes~\cite{3dioumatch2021cvpr,sess2020cvpr}. For feature-level Flip, we place feature information to its symmetric position on the feature map. For feature-level RS, we nullify randomly selected 5\% of feature data. Combination of feature-level Flip and RS is actually a feature-level variant of the SOTA 3DIoUMatch~\cite{3dioumatch2021cvpr}, named \textbf{F-3DIoUMatch}.\footnote{Policy 4$\ast$ indicates F-3DIoUMatch.}
\textbf{Noise} is an existing feature augmentation method that adds Gaussian noise, which is used for domain generalization of image classification~\cite{li2021simple}.
Lastly, \textbf{Rotation} rotates the feature with a degree randomly selected from [-45$^\circ$, 45$^\circ$] and performs bilinear interpolation.

\vspace{3pt}
\noindent
\textbf{Result analysis.}
Table~\ref{tab:feature_aug_performance} shows each augmentation scheme's performance margin compared to Baseline in the partial-label scenario.
Flip significantly underperforms Baseline despite the use of much more (unlabeled) training data, verifying that feature-level Flip damages important information in 3D scenes.
Both Noise and RS have marginal impact on performance, showing that these perturbation strategies do not result in meaningful data diversity.
Combining Flip and RS (\ie, F-3DIoUMatch) still performs worse than Baseline due to the negative effect of Flip, which confirms that naïve application of SOTA SSL methods at a feature level does not work.
Although Rotation improves performance, our \fgt provides the lowest augmentation errors (\fig\ref{subfig:error_bar}) and thus \textit{the best performance} in all cases.

\begin{figure}[t]
  \centering
  \begin{subfigure}[t]{.49\columnwidth}
    \centering
    \includegraphics[width=.9\columnwidth]{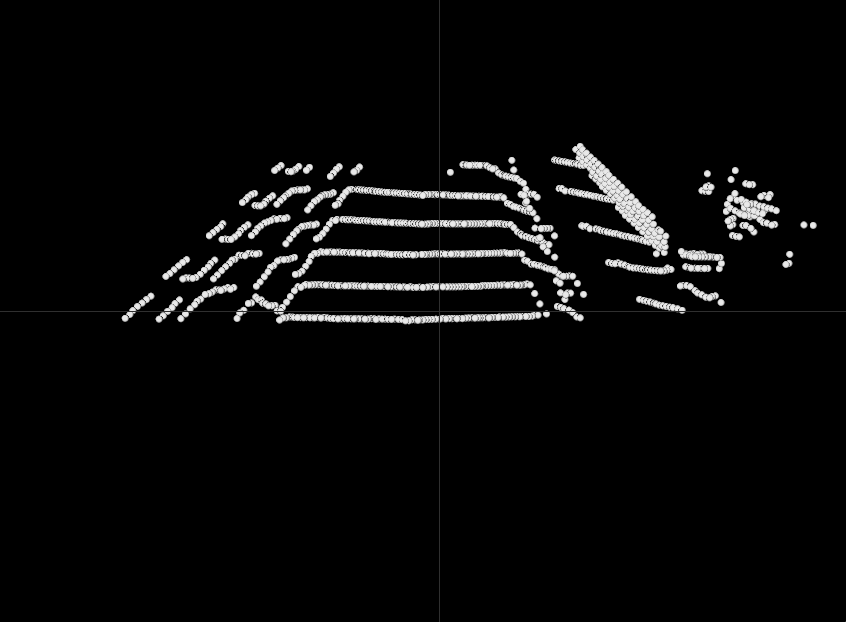}
    \caption{Original raw-point scene}
  \end{subfigure}
  \hfill
  \begin{subfigure}[t]{.49\columnwidth}
    \centering
    \includegraphics[width=.9\columnwidth]{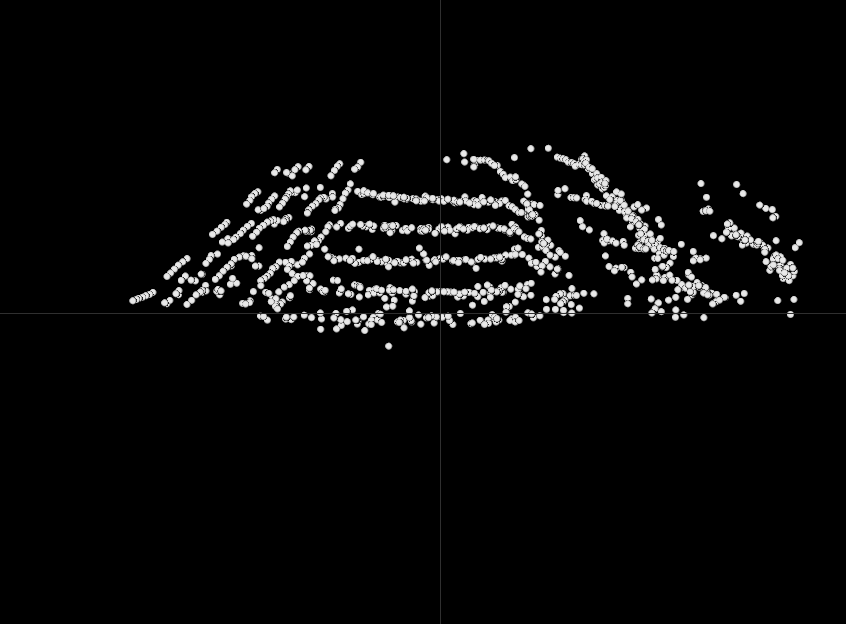}
    \caption{Restoration from the 1st layer}
  \end{subfigure}
  \hfill
  \vspace{5pt}
  \begin{subfigure}[t]{.49\columnwidth}
    \centering
    \includegraphics[width=.9\columnwidth]{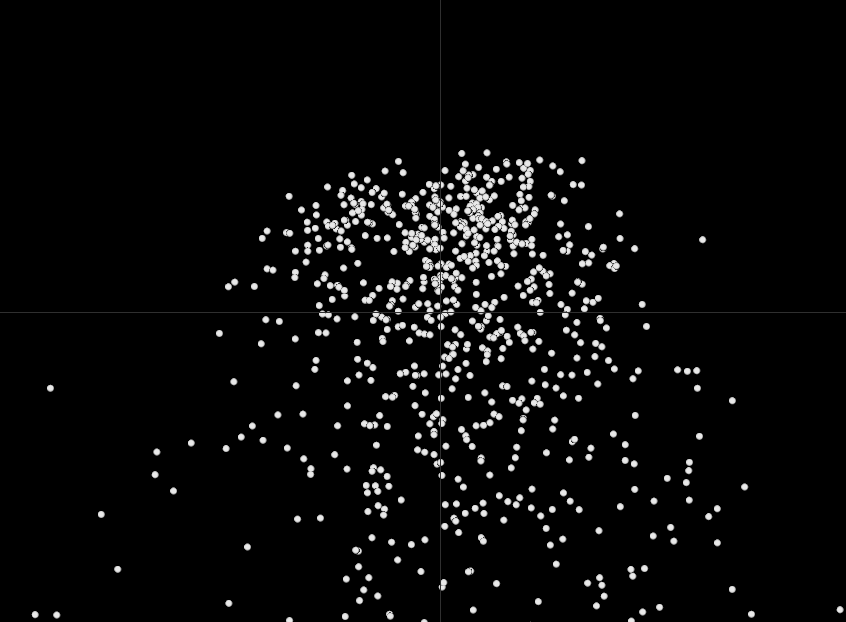}
    \caption{Restoration from a middle (3rd) layer}
  \end{subfigure}
  \hfill
  \begin{subfigure}[t]{.49\columnwidth}
    \centering
    \includegraphics[width=.9\columnwidth]{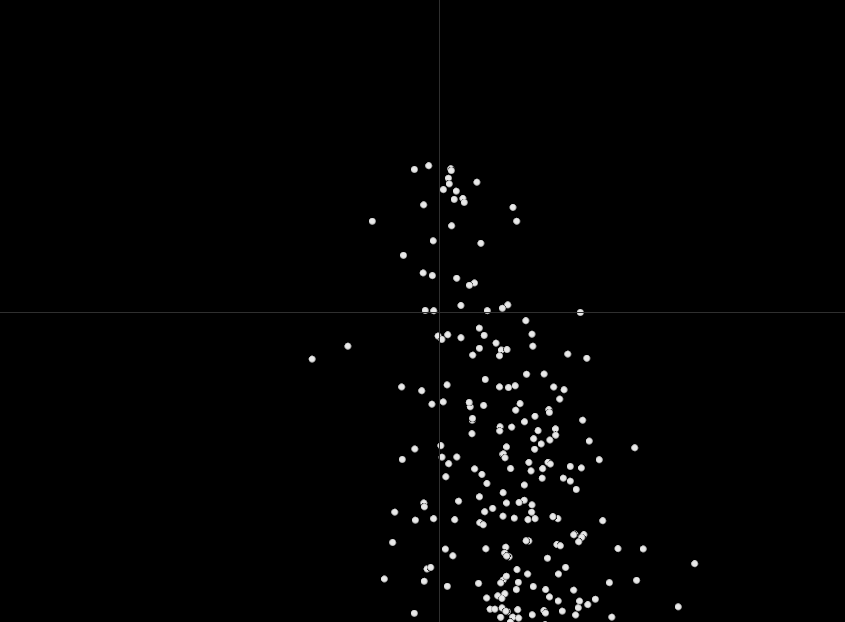}
    \caption{Restoration from the last (5th) layer, same as \uc}
  \end{subfigure}
  \hfill
  \vspace{-7pt}
  \caption{Results of inversion attack for the 3D backbone model (5 convolutional layers) of SECOND-IoU and PV-RCNN. The example 3D point cloud scene is in KITTI.}
  \label{fig:privacy}
  \vspace{-12pt}
\end{figure}


\subsection{Privacy Protection of Feature Sharing}\label{subsec:privacy}

As neural network activations could be inverted to reconstruct input data~\cite{mahendran2015understanding,dosovitskiy2016inverting,ulyanov2018deep}, there could be concerns on potential privacy leaks when sharing features.
We investigate whether an inversion attack can recover the grid-type feature data generated from both the SECOND-IoU and PV-RCNN backbone networks to the original point cloud.
To this end, we implement the inversion attack model using the decoder method~\cite{Dosovitskiy_2016_CVPR} that is widely used to evaluate whether a model consisting of convolutional layers can be inverted~\cite{NIPS2016_371bce7d,Zhao_2021_ICCV}.\footnote{To the best of our knowledge, there has been no research that particularly focuses on inversion attacks for 3D point clouds.} More details are in the supplementary material.

\vspace{3pt}
\noindent
\textbf{Result analysis.}
We conduct an inversion attack on the 3D backbone network in SECOND-IoU and PV-RCNN.\footnote{The 3D point cloud scene in \fig\ref{fig:privacy}(a) is from KITTI dataset, and the point cloud range covers the x, y, and z-axis ranges 17.6, 20, and 4 meters.}
\figs\ref{fig:privacy}(b)-(d) present the restoration results for intermediate features at three different convolutional layers of the backbone network: 1st, 3rd, and 5th (last) layers, respectively.
While the restored point cloud from the first layer is relatively similar to the original scene (\fig\ref{fig:privacy}(b)), it becomes significantly different when applied to deeper layers' features (\figs\ref{fig:privacy}(c) and (d)).
As the number of nonlinear layers increases, it becomes more difficult to accurately restore the original data. Furthermore, restoring a point cloud from its intermediate feature is particularly challenging since each raw point needs to be positioned precisely in voxelized spaces.
\uc utilizes unlabeled features at the last (deepest) layer, making it impossible to accurately recover the original scene from an intermediate feature.
Supplementary material contains more inversion examples.

\begin{table}[t]

  \caption{
    Domain adaptation results with two target datasets: KITTI and Lyft.
    Difficulty of the KITTI test dataset is set as Moderate.
    Baseline is a pre-trained model with Waymo
    whereas Oracle is trained with fully labeled target dataset.}	\label{tab:da}
   \vspace{-7pt}
  \centering
  \scalebox{.805}{

    \begin{tabular}{|c|c||c|c|}
      \hline\xrowht{5pt}
      \multirow{2}{*}{Dataset} & \multirow{2}{*}{Method}             & \multicolumn{1}{c|}{SECOND-IoU}                                              & \multicolumn{1}{c|}{PV-RCNN}                                                 \\ \cline{3-4} \xrowht{5pt}
                               &                                     & \multicolumn{1}{l|}{${\rm AP}_{\rm BEV}$ / ${\rm AP}_{\rm 3D}$}              & \multicolumn{1}{l|}{${\rm AP}_{\rm BEV}$ / ${\rm AP}_{\rm 3D}$}              \\ \hline\hline\xrowht{5pt}
      \multirow{7}{*}{Lyft}    & Baseline                            & \multicolumn{1}{c|}{30.20 / 21.32}                                           & \multicolumn{1}{c|}{33.00 / 24.49}                                           \\ \cline{2-4} \xrowht{5pt}
                               & SN                                  & \multicolumn{1}{c|}{28.38 / 19.25}                                           & \multicolumn{1}{c|}{33.44 / 25.64}                                           \\ \cline{2-4} \xrowht{5pt}
                               & ST3D                                & \multicolumn{1}{c|}{60.53 / 29.90}                                           & \multicolumn{1}{c|}{62.28 / 42.63}                                           \\ \cline{2-4} \xrowht{5pt}
                               & \cellcolor[HTML]{EFEFEF} \uc        & \multicolumn{1}{c|}{\cellcolor[HTML]{EFEFEF}\textbf{68.83} / \textbf{45.66}} & \multicolumn{1}{c|}{\cellcolor[HTML]{EFEFEF}\textbf{63.38} / \textbf{46.83}} \\ \cline{2-4} \xrowht{5pt}
                               & ST3D (w/ SN)                        & \multicolumn{1}{c|}{52.86 / 21.25}                                           & \multicolumn{1}{c|}{60.15 / 44.02}                                           \\ \cline{2-4} \xrowht{5pt}
                               & \cellcolor[HTML]{EFEFEF}\uc (w/ SN) & \multicolumn{1}{c|}{\cellcolor[HTML]{EFEFEF}\textbf{65.10} / \textbf{49.24}} & \multicolumn{1}{c|}{\cellcolor[HTML]{EFEFEF}\textbf{63.58} / \textbf{49.35}} \\ \cline{2-4} \xrowht{5pt}
                               & Oracle                              & \multicolumn{1}{c|}{76.70 / 61.70}                                           & \multicolumn{1}{c|}{78.68 / 64.54}                                           \\ \hline\hline\xrowht{5pt}
      \multirow{7}{*}{KITTI}   & Baseline                            & \multicolumn{1}{c|}{54.14 / 10.16}                                           & \multicolumn{1}{c|}{62.24 / 9.24}                                            \\ \cline{2-4}\xrowht{5pt}
                               & SN                                  & \multicolumn{1}{c|}{60.80 / 37.30}                                           & \multicolumn{1}{c|}{60.08 / \textbf{38.86}}                                  \\ \cline{2-4} \xrowht{5pt}
                               & ST3D                                & \multicolumn{1}{c|}{\textbf{70.90} / \textbf{40.16}}                         & \multicolumn{1}{c|}{\textbf{66.19} / 23.26}                                  \\ \cline{2-4} \xrowht{5pt}
                               & \cellcolor[HTML]{EFEFEF}\uc         & \multicolumn{1}{c|}{\cellcolor[HTML]{EFEFEF}58.26 / 11.71}                   & \multicolumn{1}{c|}{\cellcolor[HTML]{EFEFEF}62.09 / 11.35}                   \\ \cline{2-4} \xrowht{5pt}
                               & ST3D (w/ SN)                        & \multicolumn{1}{c|}{80.97 / 57.68}                                           & \multicolumn{1}{c|}{54.30 / 48.79}                                           \\ \cline{2-4} \xrowht{5pt}
                               & \cellcolor[HTML]{EFEFEF}\uc (w/ SN) & \multicolumn{1}{c|}{\cellcolor[HTML]{EFEFEF}\textbf{84.12} / \textbf{67.65}} & \multicolumn{1}{c|}{\cellcolor[HTML]{EFEFEF}\textbf{85.90} / \textbf{61.12}} \\ \cline{2-4} \xrowht{5pt}
                               & Oracle                              & \multicolumn{1}{c|}{90.36 / 82.02}                                           & \multicolumn{1}{c|}{90.84 / 84.56}                                           \\ \hline
    \end{tabular}
  }
  \vspace{-7pt}
\end{table}

\begin{figure*}[t]
  \begin{minipage}{.65\linewidth}

    \captionof{table}{
      Partial-label scenario results with three portions of labeled data in the KITTI dataset: 2\%, 10\%, 25\%.}
    \label{tab:partial}
    \vspace{-4pt}
    \centering
    \scalebox{.7}{
      \begin{tabular}[b]{|cc|ccc|ccc|ccc|}
        \hline
        \multicolumn{2}{|c|}{\multirow{2}{*}{${\rm AP}_{\rm 3D}$}} & \multicolumn{3}{c|}{2\%}              & \multicolumn{3}{c|}{10\%}                                   & \multicolumn{3}{c|}{25\%}                                                                                                                                                                                                                                                                                                                                                                                                                \\ \cline{3-11}
        \multicolumn{2}{|l|}{}                                     & \multicolumn{1}{l|}{\small Easy}      & \multicolumn{1}{l|}{\small Mod}                             & {\small Hard}                                               & \multicolumn{1}{l|}{\small Easy}       & \multicolumn{1}{l|}{\small Mod}                            & {\small Hard}                                              & \multicolumn{1}{l|}{\small Easy}      & \multicolumn{1}{l|}{\small Mod}                            & {\small Hard}                                                                                      \\ \hline
        \multicolumn{1}{|l|}{\multirow{5}{*}{SECOND-IoU}}          & Baseline                              & \multicolumn{1}{l|}{56.69}                                  & \multicolumn{1}{l|}{44.11}                                  & 37.19                                  & \multicolumn{1}{l|}{70.58}                                 & \multicolumn{1}{l|}{56.00}                                 & 47.94                                 & \multicolumn{1}{l|}{84.47}                                 & \multicolumn{1}{l|}{71.06}                                 & 62.87                                 \\ \cline{2-11}
        \multicolumn{1}{|l|}{}                                     & 3DIoUMatch                            & \multicolumn{1}{l|}{63.57}                                  & \multicolumn{1}{l|}{49.58}                                  & 43.00                                  & \multicolumn{1}{l|}{71.76}                                 & \multicolumn{1}{l|}{57.01}                                 & 50.08                                 & \multicolumn{1}{l|}{81.71}                                 & \multicolumn{1}{l|}{68.51}                                 & 60.92                                 \\ \cline{2-11}
        \multicolumn{1}{|l|}{}                                     & \cellcolor[HTML]{EFEFEF}improved (\%) & \multicolumn{1}{l|}{\cellcolor[HTML]{EFEFEF}12.13}          & \multicolumn{1}{l|}{\cellcolor[HTML]{EFEFEF}12.39}          & \cellcolor[HTML]{EFEFEF}15.62          & \multicolumn{1}{l|}{\cellcolor[HTML]{EFEFEF}1.67}          & \multicolumn{1}{l|}{\cellcolor[HTML]{EFEFEF}1.80}          & \cellcolor[HTML]{EFEFEF}4.47          & \multicolumn{1}{l|}{\cellcolor[HTML]{EFEFEF}-3.26}         & \multicolumn{1}{l|}{\cellcolor[HTML]{EFEFEF}-3.59}         & \cellcolor[HTML]{EFEFEF}-3.11         \\ \cline{2-11}
        \multicolumn{1}{|l|}{}                                     & \uc                                   & \multicolumn{1}{l|}{70.19}                                  & \multicolumn{1}{l|}{59.97}                                  & 44.83                                  & \multicolumn{1}{l|}{76.09}                                 & \multicolumn{1}{l|}{60.41}                                 & 51.84                                 & \multicolumn{1}{l|}{85.22}                                 & \multicolumn{1}{l|}{72.87}                                 & 63.93                                 \\ \cline{2-11}
        \multicolumn{1}{|l|}{}                                     & \cellcolor[HTML]{EFEFEF}improved (\%) & \multicolumn{1}{l|}{\cellcolor[HTML]{EFEFEF}\textbf{23.81}} & \multicolumn{1}{l|}{\cellcolor[HTML]{EFEFEF}\textbf{35.96}} & \cellcolor[HTML]{EFEFEF}\textbf{20.54} & \multicolumn{1}{l|}{\cellcolor[HTML]{EFEFEF}\textbf{7.81}} & \multicolumn{1}{l|}{\cellcolor[HTML]{EFEFEF}\textbf{7.87}} & \cellcolor[HTML]{EFEFEF}\textbf{8.14} & \multicolumn{1}{l|}{\cellcolor[HTML]{EFEFEF}\textbf{0.89}} & \multicolumn{1}{l|}{\cellcolor[HTML]{EFEFEF}\textbf{2.55}} & \cellcolor[HTML]{EFEFEF}\textbf{1.69} \\ \hline\hline
        \multicolumn{1}{|c|}{\multirow{5}{*}{PV-RCNN}}             & Baseline                              & \multicolumn{1}{l|}{68.10}                                  & \multicolumn{1}{l|}{53.27}                                  & 46.20                                  & \multicolumn{1}{l|}{81.23}                                 & \multicolumn{1}{l|}{68.67}                                 & 60.32                                 & \multicolumn{1}{l|}{87.63}                                 & \multicolumn{1}{l|}{76.03}                                 & 68.62                                 \\ \cline{2-11}
        \multicolumn{1}{|l|}{}                                     & 3DIoUMatch                            & \multicolumn{1}{l|}{81.04}                                  & \multicolumn{1}{l|}{65.77}                                  & 58.83                                  & \multicolumn{1}{l|}{85.26}                                 & \multicolumn{1}{l|}{70.64}                                 & 63.32                                 & \multicolumn{1}{l|}{85.08}                                 & \multicolumn{1}{l|}{72.37}                                 & 65.02                                 \\ \cline{2-11}
        \multicolumn{1}{|l|}{}                                     & \cellcolor[HTML]{EFEFEF}improved (\%) & \multicolumn{1}{l|}{\cellcolor[HTML]{EFEFEF}\textbf{19.00}} & \multicolumn{1}{l|}{\cellcolor[HTML]{EFEFEF}\textbf{23.47}} & \cellcolor[HTML]{EFEFEF}\textbf{27.34} & \multicolumn{1}{l|}{\cellcolor[HTML]{EFEFEF}\textbf{4.97}} & \multicolumn{1}{l|}{\cellcolor[HTML]{EFEFEF}\textbf{2.87}} & \cellcolor[HTML]{EFEFEF}4.98          & \multicolumn{1}{l|}{\cellcolor[HTML]{EFEFEF}-2.91}         & \multicolumn{1}{l|}{\cellcolor[HTML]{EFEFEF}-4.81}         & \cellcolor[HTML]{EFEFEF}-5.25         \\ \cline{2-11}
        \multicolumn{1}{|l|}{}                                     & \uc                                   & \multicolumn{1}{l|}{76.46}                                  & \multicolumn{1}{l|}{61.44}                                  & 52.94                                  & \multicolumn{1}{l|}{83.64}                                 & \multicolumn{1}{l|}{69.60}                                 & 63.53                                 & \multicolumn{1}{l|}{88.05}                                 & \multicolumn{1}{l|}{76.61}                                 & 70.80                                 \\ \cline{2-11}
        \multicolumn{1}{|l|}{}                                     & \cellcolor[HTML]{EFEFEF}improved (\%) & \multicolumn{1}{l|}{\cellcolor[HTML]{EFEFEF}12.28}          & \multicolumn{1}{l|}{\cellcolor[HTML]{EFEFEF}15.34}          & \cellcolor[HTML]{EFEFEF}14.59          & \multicolumn{1}{l|}{\cellcolor[HTML]{EFEFEF}2.97}          & \multicolumn{1}{l|}{\cellcolor[HTML]{EFEFEF}1.35}          & \cellcolor[HTML]{EFEFEF}\textbf{5.32} & \multicolumn{1}{l|}{\cellcolor[HTML]{EFEFEF}\textbf{0.48}} & \multicolumn{1}{l|}{\cellcolor[HTML]{EFEFEF}\textbf{0.76}} & \cellcolor[HTML]{EFEFEF}\textbf{3.18} \\
        \hline
      \end{tabular}
    }
  \end{minipage}
  \begin{minipage}{.35\linewidth}
  \vspace{-18pt}
    \centering
    \includegraphics[width=0.85\columnwidth]{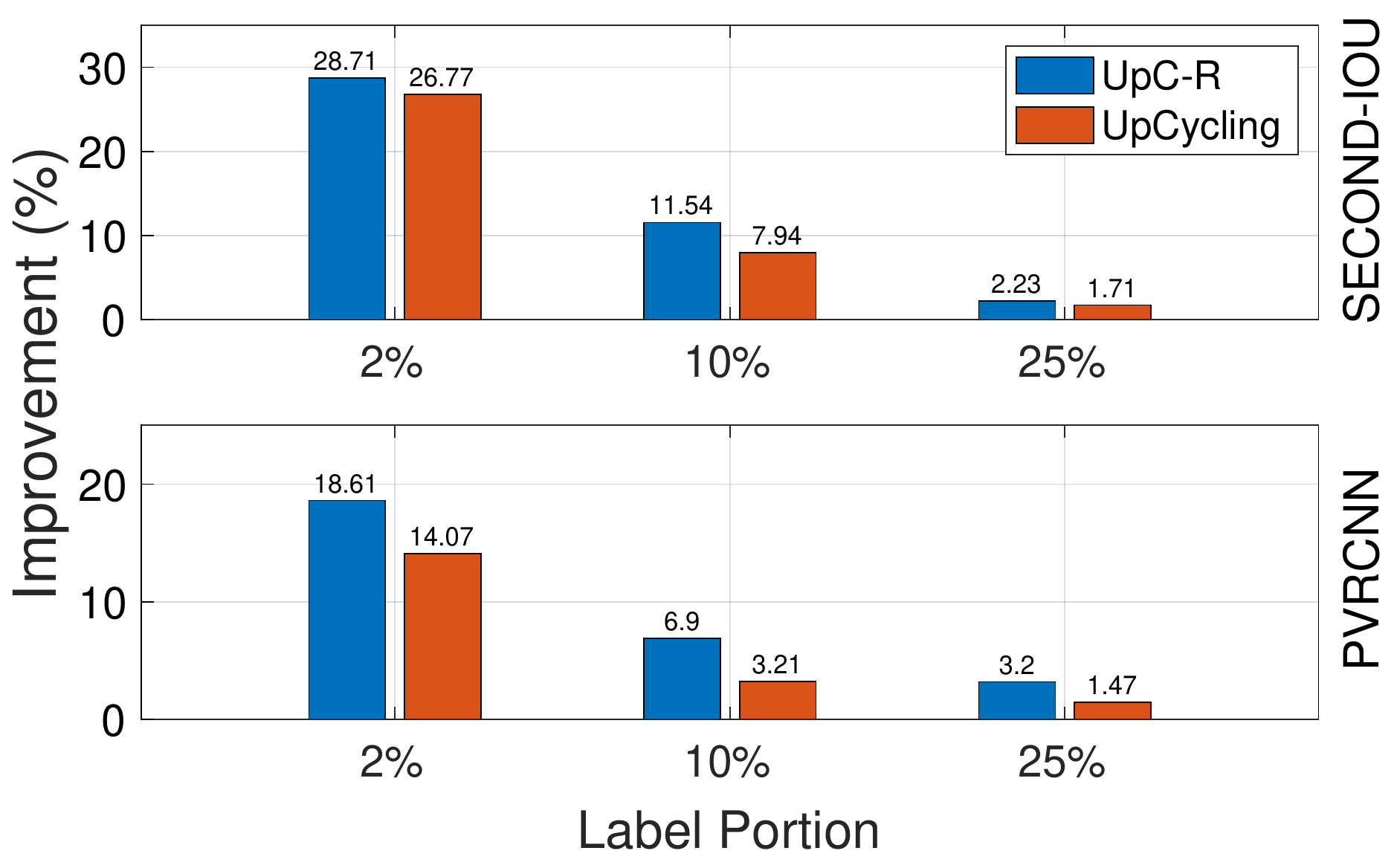}
    \vspace{-10pt}
    \caption{\ucr vs. \uc: Partial-label results in the KITTI dataset. The average performance improvement in all KITTI test cases (easy, moderate, and hard).} \label{fig:upcrvs.upc}
    \vspace{-22pt}
  \end{minipage}
  \vspace{-10pt}
\end{figure*}

\subsection{Domain Adaptation Experiments} \label{subsec:domain_adaptation}

Although \uc offers privacy protection by using only intermediate features, it is crucial to evaluate whether it provides competitive detection accuracy compared to the SOTA methods that use raw-level point clouds (Sections~\ref{subsec:domain_adaptation} and \ref{subsec:partial-label}). In domain adaptation experiments, we use the Waymo dataset as the source domain and the Lyft and KITTI datasets as the target domains. The model is first pre-trained on the source domain's labeled data (called the baseline model), adapted using unlabeled training data in a target domain, and then tested on the target domain's test data.

\vspace{3pt}
\noindent
\textbf{Comparison schemes.}
We compare \uc with various methods. \textbf{Baseline} evaluates the baseline model directly and \textbf{Oracle} adapts the model with fully supervised learning in the target domain, which provide the lower- and upper-bound performance, respectively.
\textbf{ST3D}~\cite{st3d2021cvpr} and \textbf{SN} (Statistical Normalization)~\cite{wang2020train} are the SOTA domain adaptation methods on 3D object detection that utilize unlabeled raw 3D scenes.
ST3D generates pseudo labels from unlabeled data in the target domain to adapt the baseline model. SN assumes that statistical object sizes in the target domain are given and trains the baseline model in the source domain using the target domain object size information.
We also evaluate variants of ST3D and our \uc by combining SN together, denoted as \textbf{(w/ SN)}.

\vspace{3pt}
\noindent
\textbf{Result analysis.}
Table~\ref{tab:da} shows the results of \uc and the various comparison methods on SECOND-IoU and PV-RCNN. Surprisingly, the results show that although \uc (or w/ SN) does not utilize raw-point scenes for privacy protection, it \textit{provides the best accuracy} in most cases. Specifically, \uc (or w/ SN) significantly outperforms the two SOTA methods (ST3D and SN) in the Lyft case.
When compared to the better option between ST3D (or w/ SN) and SN in each case, \uc improves accuracy by \textbf{1.3$\sim$19.71} ${\rm AP}_{\rm BEV}$ and \textbf{5.33$\sim$19.34} ${\rm AP}_{\rm 3D}$.
The results demonstrate the effectiveness of hybrid pseudo labels and feature-level augmentation schemes in \uc and also suggest the potential of using unlabeled features to advance 3D object detection models.

Taking a deeper look, SN significantly improves \uc performance in the KITTI dataset. Since object sizes in KITTI are different from those in Lyft and Waymo, adjusting object sizes with SN for \uc is effective.
%

\subsection{Partial-label Experiments} \label{subsec:partial-label}

In partial-label experiments, we use the same setting as in Section~\ref{subsec:featureaug} but train both SECOND-IoU and PV-RCNN. %

\vspace{3pt}
\noindent
\textbf{Comparison schemes.}
In this scenario, \textbf{Baseline} trains the model using only the limited amount of labeled data.
\textbf{3DIoUMatch}~\cite{3dioumatch2021cvpr} is the SOTA SSL method using unlabeled raw-point scenes. For consistency regularization, 3DIoUMatch uses Flip and RS to augment raw data and filters pseudo labels in the IoU-guided NMS.\footnote{Since the authors in~\cite{3dioumatch2021cvpr} did not use the rearranged KITTI dataset in their experiments, we measure the performance of 3DIoUMatch again in the rearranged KITTI dataset. In addition, we newly implement 3DIoUMatch on SECOND-IoU for more extensive comparison.
}

\vspace{3pt}
\noindent
\textbf{Result analysis.}
Table~\ref{tab:partial} shows that \uc outperforms 3DIoUMatch in most cases by effectively utilizing unlabeled feature-level data. In the case of 25\%, 3DIoUMatch even underperforms Baseline but \uc maintains performance improvement on both SECOND-IoU and PV-RCNN. 
The results are interesting because the scenario is unfavorable for \uc in that (1) \uc trains the 3D backbone only using the small portion of labeled data and (2) the effect of \fgt could be marginal since the number of GT samples are proportional to that of labeled data.
\uc successfully overcomes the disadvantages, verifying that it achieves significant performance improvement even when using a relatively immature backbone network and \fgt effectively augments a large number of unlabeled data when only a small number of GTs are available.  


\subsection{Ablation Studies}\label{sec:discussion}
\vspace{-2pt}
Since \uc freezes the backbone during the SSL process for effective feature sharing, we evaluate the effect of the backbone freezing. To this end, we devise a comparison scheme \textbf{\ucr},  the application of \uc at the raw-level input. \ucr augments a raw-level 3D scene using GT samples and trains the whole network including the backbone using unlabeled data and hybrid pseudo labels. Note that this approach not only sacrifices privacy but also takes much longer to train compared to \uc.

\vspace{1.5pt}
\noindent
\textbf{Result analysis.}
\fig\ref{fig:upcrvs.upc} compares \ucr and \uc in the partial-label scenario in Section~\ref{subsec:partial-label}. While sacrificing privacy, \ucr outperforms \uc by training the backbone further. Interestingly, \ucr performs even better than the SOTA 3DIoUMatch (Table~\ref{tab:partial}), demonstrating that GT sampling is more effective augmentation than the combination of Flip and RS \textit{even at the raw-input level}.
On the other hand, the performance gap between \ucr and \uc decreases as the number of labeled data increases, meaning that once the backbone is well-trained, the combination of hybrid pseudo-labels and GT-based augmentation can be applied flexibly to any layer without performance degradation. We see this as the unique advantage of GT sampling that other point cloud augmentation methods cannot provide.  

\vspace{-2pt}
\section{Conclusion}
\label{sec:conclusion}
\vspace{-1pt}

In this paper, we present that \uc, a novel semi-supervised learning method for \ddd models, improves model performance by gathering de-identified unlabeled data from \avs.
To the best of our knowledge, no study has considered labeling cost, privacy, and edge computing resources in \avs and overall systems altogether.
Taking all these factors into account, we apply \uc to the representative \ddd models, SECOND-IoU and PV-RCNN.
Through various experiments using multiple datasets, we verify the superiority of \uc in partial-label situations as well as domain adaptation in comparison with other SOTA methods.
Furthermore, we also confirm that feature-level GT sampling can improve model performance significantly compared with other augmentation methods applied at a feature level.


\begin{thebibliography}{10}\itemsep=-1pt

\bibitem{kittisorting}
bostondiditeam.
\newblock Exploratory findings for the kitti vision benchmark suite.
\newblock \url{https://github.com/bostondiditeam}, 2017.

\bibitem{cen2021deep}
Feng Cen, Xiaoyu Zhao, Wuzhuang Li, and Guanghui Wang.
\newblock Deep feature augmentation for occluded image classification.
\newblock {\em Pattern Recognition}, 111:107737, 2021.

\bibitem{chu2020feature}
Peng Chu, Xiao Bian, Shaopeng Liu, and Haibin Ling.
\newblock Feature space augmentation for long-tailed data.
\newblock In {\em European Conference on Computer Vision}, pages 694--710.
  Springer, 2020.

\bibitem{cubuk2020randaugment}
Ekin~D Cubuk, Barret Zoph, Jonathon Shlens, and Quoc~V Le.
\newblock Randaugment: Practical automated data augmentation with a reduced
  search space.
\newblock In {\em Proceedings of the IEEE/CVF Conference on Computer Vision and
  Pattern Recognition Workshops}, pages 702--703, 2020.

\bibitem{voxelrcnn2021aaai}
Jiajun Deng, Shaoshuai Shi, Peiwei Li, W. Zhou, Yanyong Zhang, and H. Li.
\newblock Voxel r-cnn: Towards high performance voxel-based 3d object
  detection.
\newblock In {\em AAAI}, 2021.

\bibitem{devries2017improved}
Terrance DeVries and Graham~W Taylor.
\newblock Improved regularization of convolutional neural networks with cutout.
\newblock {\em arXiv preprint arXiv:1708.04552}, 2017.

\bibitem{NIPS2016_371bce7d}
Alexey Dosovitskiy and Thomas Brox.
\newblock Generating images with perceptual similarity metrics based on deep
  networks.
\newblock In D. Lee, M. Sugiyama, U. Luxburg, I. Guyon, and R. Garnett,
  editors, {\em Advances in Neural Information Processing Systems}, volume~29.
  Curran Associates, Inc., 2016.

\bibitem{Dosovitskiy_2016_CVPR}
Alexey Dosovitskiy and Thomas Brox.
\newblock Inverting visual representations with convolutional networks.
\newblock In {\em Proceedings of the IEEE Conference on Computer Vision and
  Pattern Recognition (CVPR)}, June 2016.

\bibitem{dosovitskiy2016inverting}
Alexey Dosovitskiy and Thomas Brox.
\newblock Inverting visual representations with convolutional networks.
\newblock In {\em Proceedings of the IEEE conference on computer vision and
  pattern recognition}, pages 4829--4837, 2016.

\bibitem{privacy14secpri}
David Eckhoff and Christoph Sommer.
\newblock Driving for big data? privacy concerns in vehicular networking.
\newblock {\em Security \& Privacy, IEEE}, 12:77--79, 01 2014.

\bibitem{kitti2012CVPR}
Andreas Geiger, Philip Lenz, and Raquel Urtasun.
\newblock Are we ready for autonomous driving? the kitti vision benchmark
  suite.
\newblock In {\em Conference on Computer Vision and Pattern Recognition
  (CVPR)}, 2012.

\bibitem{gupta2018distributed}
Otkrist Gupta and Ramesh Raskar.
\newblock Distributed learning of deep neural network over multiple agents.
\newblock {\em Journal of Network and Computer Applications}, 116:1--8, 2018.

\bibitem{hahner2020quantifying}
Martin Hahner, Dengxin Dai, Alexander Liniger, and Luc Van~Gool.
\newblock Quantifying data augmentation for lidar based 3d object detection.
\newblock {\em arXiv preprint arXiv:2004.01643}, 2020.

\bibitem{lyft_dataset}
John Houston, Guido Zuidhof, Luca Bergamini, Yawei Ye, Ashesh Jain, Sammy
  Omari, Vladimir Iglovikov, and Peter Ondruska.
\newblock One thousand and one hours: Self-driving motion prediction dataset.
\newblock {\em CoRR}, abs/2006.14480, 2020.

\bibitem{csd19nips}
Jisoo Jeong, Seungeui Lee, Jeesoo Kim, and Nojun Kwak.
\newblock Consistency-based semi-supervised learning for object detection.
\newblock In H. Wallach, H. Larochelle, A. Beygelzimer, F. d~\textquotesingle
  Alch\'{e}-Buc, E. Fox, and R. Garnett, editors, {\em Advances in Neural
  Information Processing Systems 32}, pages 10759--10768. Curran Associates,
  Inc., 2019.

\bibitem{Konecny2017arxiv}
Jakub Konečný, H.~Brendan McMahan, Felix~X. Yu, Peter Richtárik,
  Ananda~Theertha Suresh, and Dave Bacon.
\newblock Federated learning: Strategies for improving communication
  efficiency, 2016.

\bibitem{kumar2019closer}
Varun Kumar, Hadrien Glaude, Cyprien de Lichy, and William Campbell.
\newblock A closer look at feature space data augmentation for few-shot intent
  classification.
\newblock {\em arXiv preprint arXiv:1910.04176}, 2019.

\bibitem{Laine2017iclr}
Samuli Laine and Timo Aila.
\newblock Temporal ensembling for semi-supervised learning, 2016.

\bibitem{pointpillar2019cvpr}
A.~H. {Lang}, S. {Vora}, H. {Caesar}, L. {Zhou}, J. { Yang}, and O. {Beijbom}.
\newblock Pointpillars: Fast encoders for object detection from point clouds.
\newblock In {\em 2019 IEEE/CVF Conference on Computer Vision and Pattern
  Recognition (CVPR)}, pages 12689--12697, June 2019.

\bibitem{lee2013pseudo}
Dong-Hyun Lee.
\newblock Pseudo-label : The simple and efficient semi-supervised learning
  method for deep neural networks.
\newblock {\em ICML 2013 Workshop : Challenges in Representation Learning
  (WREPL )}, 07 2013.

\bibitem{li2021simple}
Pan Li, Da Li, Wei Li, Shaogang Gong, Yanwei Fu, and Timothy~M Hospedales.
\newblock A simple feature augmentation for domain generalization.
\newblock In {\em Proceedings of the IEEE/CVF International Conference on
  Computer Vision}, pages 8886--8895, 2021.

\bibitem{Liu_2018_CVPR}
Bo Liu, Xudong Wang, Mandar Dixit, Roland Kwitt, and Nuno Vasconcelos.
\newblock Feature space transfer for data augmentation.
\newblock In {\em Proceedings of the IEEE Conference on Computer Vision and
  Pattern Recognition (CVPR)}, June 2018.

\bibitem{mahendran2015understanding}
Aravindh Mahendran and Andrea Vedaldi.
\newblock Understanding deep image representations by inverting them.
\newblock In {\em Proceedings of the IEEE conference on computer vision and
  pattern recognition}, pages 5188--5196, 2015.

\bibitem{fl2017pmlr}
Brendan McMahan, Eider Moore, Daniel Ramage, Seth Hampson, and Blaise~Aguera y
  Arcas.
\newblock {Communication-Efficient Learning of Deep Networks from Decentralized
  Data}.
\newblock In Aarti Singh and Jerry Zhu, editors, {\em Proceedings of the 20th
  International Conference on Artificial Intelligence and Statistics},
  volume~54 of {\em Proceedings of Machine Learning Research}, pages
  1273--1282, Fort Lauderdale, FL, USA, 20--22 Apr 2017. PMLR.

\bibitem{e_pp_v2x2020sensors}
Yang Ming and Xiaopeng Yu.
\newblock Efficient privacy-preserving data sharing for fog-assisted vehicular
  sensor networks.
\newblock {\em Sensors}, 20(2), 2020.

\bibitem{Miyato2019tpami}
T. {Miyato}, S. {Maeda}, M. {Koyama}, and S. {Ishii}.
\newblock Virtual adversarial training: A regularization method for supervised
  and semi-supervised learning.
\newblock {\em IEEE Transactions on Pattern Analysis and Machine Intelligence},
  41(8):1979--1993, 2019.

\bibitem{pointnet2017cvpr}
Charles~R Qi, Hao Su, Kaichun Mo, and Leonidas~J Guibas.
\newblock Pointnet: Deep learning on point sets for 3d classification and
  segmentation.
\newblock In {\em Proceedings of the IEEE conference on computer vision and
  pattern recognition}, pages 652--660, 2017.

\bibitem{qi2017pointnet++}
Charles~Ruizhongtai Qi, Li Yi, Hao Su, and Leonidas~J Guibas.
\newblock Pointnet++: Deep hierarchical feature learning on point sets in a
  metric space.
\newblock {\em Advances in neural information processing systems}, 30, 2017.

\bibitem{reddi2020adaptive}
Sashank Reddi, Zachary Charles, Manzil Zaheer, Zachary Garrett, Keith Rush,
  Jakub Kone{\v{c}}n{\`y}, Sanjiv Kumar, and H~Brendan McMahan.
\newblock Adaptive federated optimization.
\newblock {\em arXiv preprint arXiv:2003.00295}, 2020.

\bibitem{fedpaq2020pmlr}
Amirhossein Reisizadeh, Aryan Mokhtari, Hamed Hassani, Ali Jadbabaie, and
  Ramtin Pedarsani.
\newblock Fedpaq: A communication-efficient federated learning method with
  periodic averaging and quantization.
\newblock In Silvia Chiappa and Roberto Calandra, editors, {\em Proceedings of
  the Twenty Third International Conference on Artificial Intelligence and
  Statistics}, volume 108 of {\em Proceedings of Machine Learning Research},
  pages 2021--2031. PMLR, 26--28 Aug 2020.

\bibitem{pvrcnn2020cvpr}
Shaoshuai Shi, Chaoxu Guo, Li Jiang, Zhe Wang, Jianping Shi, Xiaogang Wang, and
  Hongsheng Li.
\newblock Pv-rcnn: Point-voxel feature set abstraction for 3d object detection.
\newblock In {\em Proceedings of the IEEE/CVF Conference on Computer Vision and
  Pattern Recognition (CVPR)}, June 2020.

\bibitem{pointrcnn2019CVPR}
Shaoshuai Shi, Xiaogang Wang, and Hongsheng Li.
\newblock Pointrcnn: 3d object proposal generation and detection from point
  cloud.
\newblock In {\em The IEEE Conference on Computer Vision and Pattern
  Recognition (CVPR)}, June 2019.

\bibitem{uveqfed2021tsp}
Nir Shlezinger, Mingzhe Chen, Yonina~C. Eldar, H.~Vincent Poor, and Shuguang
  Cui.
\newblock Uveqfed: Universal vector quantization for federated learning.
\newblock {\em IEEE Transactions on Signal Processing}, 69:500--514, 2021.

\bibitem{singh2019detailed}
Abhishek Singh, Praneeth Vepakomma, Otkrist Gupta, and Ramesh Raskar.
\newblock Detailed comparison of communication efficiency of split learning and
  federated learning.
\newblock {\em arXiv preprint arXiv:1909.09145}, 2019.

\bibitem{sohn2020fixmatch}
Kihyuk Sohn, David Berthelot, Chun-Liang Li, Zizhao Zhang, Nicholas Carlini,
  Ekin~D. Cubuk, Alex Kurakin, Han Zhang, and Colin Raffel.
\newblock Fixmatch: Simplifying semi-supervised learning with consistency and
  confidence.
\newblock {\em arXiv preprint arXiv:2001.07685}, 2020.

\bibitem{waymopaper2020cvpr}
Pei Sun, Henrik Kretzschmar, Xerxes Dotiwalla, Aurelien Chouard, Vijaysai
  Patnaik, Paul Tsui, James Guo, Yin Zhou, Yuning Chai, Benjamin Caine, Vijay
  Vasudevan, Wei Han, Jiquan Ngiam, Hang Zhao, Aleksei Timofeev, Scott
  Ettinger, Maxim Krivokon, Amy Gao, Aditya Joshi, Yu Zhang, Jonathon Shlens,
  Zhifeng Chen, and Dragomir Anguelov.
\newblock Scalability in perception for autonomous driving: Waymo open dataset.
\newblock In {\em Proceedings of the IEEE/CVF Conference on Computer Vision and
  Pattern Recognition (CVPR)}, June 2020.

\bibitem{Tarvainen2017nips}
Antti Tarvainen and Harri Valpola.
\newblock Mean teachers are better role models: Weight-averaged consistency
  targets improve semi-supervised deep learning results, 2017.

\bibitem{openpcdet2020}
OpenPCDet~Development Team.
\newblock Openpcdet: An open-source toolbox for 3d object detection from point
  clouds.
\newblock \url{https://github.com/open-mmlab/OpenPCDet}, 2020.

\bibitem{ulyanov2018deep}
Dmitry Ulyanov, Andrea Vedaldi, and Victor Lempitsky.
\newblock Deep image prior.
\newblock In {\em Proceedings of the IEEE conference on computer vision and
  pattern recognition}, pages 9446--9454, 2018.

\bibitem{vepakomma2018split}
Praneeth Vepakomma, Otkrist Gupta, Tristan Swedish, and Ramesh Raskar.
\newblock Split learning for health: Distributed deep learning without sharing
  raw patient data.
\newblock {\em arXiv preprint arXiv:1812.00564}, 2018.

\bibitem{vepakomma2018no}
Praneeth Vepakomma, Tristan Swedish, Ramesh Raskar, Otkrist Gupta, and
  Abhimanyu Dubey.
\newblock No peek: A survey of private distributed deep learning.
\newblock {\em arXiv preprint arXiv:1812.03288}, 2018.

\bibitem{pmlr-v97-verma19a}
Vikas Verma, Alex Lamb, Christopher Beckham, Amir Najafi, Ioannis Mitliagkas,
  David Lopez-Paz, and Yoshua Bengio.
\newblock Manifold mixup: Better representations by interpolating hidden
  states.
\newblock In Kamalika Chaudhuri and Ruslan Salakhutdinov, editors, {\em
  Proceedings of the 36th International Conference on Machine Learning},
  volume~97 of {\em Proceedings of Machine Learning Research}, pages
  6438--6447. PMLR, 09--15 Jun 2019.

\bibitem{3dioumatch2021cvpr}
He Wang, Yezhen Cong, Or Litany, Yue Gao, and Leonidas~J Guibas.
\newblock 3dioumatch: Leveraging iou prediction for semi-supervised 3d object
  detection.
\newblock In {\em Proceedings of the IEEE/CVF Conference on Computer Vision and
  Pattern Recognition}, pages 14615--14624, 2021.

\bibitem{wang2020federated}
Hongyi Wang, Mikhail Yurochkin, Yuekai Sun, Dimitris Papailiopoulos, and
  Yasaman Khazaeni.
\newblock Federated learning with matched averaging.
\newblock {\em arXiv preprint arXiv:2002.06440}, 2020.

\bibitem{wang2020train}
Yan Wang, Xiangyu Chen, Yurong You, Li~Erran Li, Bharath Hariharan, Mark
  Campbell, Kilian~Q Weinberger, and Wei-Lun Chao.
\newblock Train in germany, test in the usa: Making 3d object detectors
  generalize.
\newblock In {\em Proceedings of the IEEE/CVF Conference on Computer Vision and
  Pattern Recognition}, pages 11713--11723, 2020.

\bibitem{xiong2020edge}
Jinbo Xiong, Renwan Bi, Mingfeng Zhao, Jingda Guo, and Qing Yang.
\newblock Edge-assisted privacy-preserving raw data sharing framework for
  connected autonomous vehicles.
\newblock {\em IEEE Wireless Communications}, 27(3):24--30, 2020.

\bibitem{second2018sensors}
Yan Yan, Yuxing Mao, and Bo Li.
\newblock Second: Sparsely embedded convolutional detection.
\newblock {\em Sensors}, 18(10):3337, Oct 2018.

\bibitem{st3d2021cvpr}
Jihan Yang, Shaoshuai Shi, Zhe Wang, Hongsheng Li, and Xiaojuan Qi.
\newblock St3d: Self-training for unsupervised domain adaptation on 3d object
  detection.
\newblock In {\em Proceedings of the IEEE/CVF Conference on Computer Vision and
  Pattern Recognition}, pages 10368--10378, 2021.

\bibitem{sess2020cvpr}
Na Zhao, Tat-Seng Chua, and Gim~Hee Lee.
\newblock Sess: Self-ensembling semi-supervised 3d object detection.
\newblock In {\em Proceedings of the IEEE/CVF Conference on Computer Vision and
  Pattern Recognition}, pages 11079--11087, 2020.

\bibitem{Zhao_2021_ICCV}
Xuejun Zhao, Wencan Zhang, Xiaokui Xiao, and Brian Lim.
\newblock Exploiting explanations for model inversion attacks.
\newblock In {\em Proceedings of the IEEE/CVF International Conference on
  Computer Vision (ICCV)}, pages 682--692, October 2021.

\bibitem{voxelnet2018cvpr}
Yin Zhou and Oncel Tuzel.
\newblock Voxelnet: End-to-end learning for point cloud based 3d object
  detection.
\newblock In {\em 2018 IEEE/CVF Conference on Computer Vision and Pattern
  Recognition}, pages 4490--4499, 2018.

\end{thebibliography}
\end{document}